\documentclass[11pt]{article}
\usepackage{graphicx}
\usepackage{amsthm}
\usepackage{url}
\usepackage{vmargin}
\usepackage[round]{natbib}
\usepackage{hyperref}
\hypersetup{pdfauthor={Nicolas Saunier and Sophie Midenet},pdftitle={Automatic Estimation of the Exposure to Lateral Collision in Signalized Intersections using Video Sensors}, pdfkeywords={road safety, exposure, automated detection, video data, intelligent transportation systems, traffic management}}

\newcommand{\keyw}[1]{{\bf #1}}
\theoremstyle{definition}
\newtheorem{myalgorithm}{Algorithm}%

\title{Automatic Estimation of the Exposure to Lateral~Collision in~Signalized Intersections using~Video~Sensors}

\author{Nicolas Saunier\footnote{\'Ecole Polytechnique de Montr\'eal, C.P. 6079, succ. Centre-Ville, Montr\'eal Qu\'ebec, Canada, H3C 3A7. Tel.: +1 (514) 340-4711 (ext. 4962). Fax: +1 (514) 340-3981. Email: nicolas.saunier@polymtl.ca. Website: \url{http://nicolas.saunier.confins.net}}\ \ 
  and Sophie Midenet\footnote{Universit\'e Paris-Est, INRETS, GRETIA, 2 rue de la Butte Verte, 93166 Noisy-le-Grand cedex, France. Email: sophie.midenet@inrets.fr}}

\begin{document}

\maketitle

\thispagestyle{empty}

\section*{Abstract:}
Intersections constitute one of the most dangerous elements in road systems. Traffic signals remain the most common way to control traffic at high-volume intersections and offer many opportunities to apply intelligent transportation systems to make traffic more efficient and safe. This paper describes an automated method to estimate the temporal exposure of road users crossing the conflict zone to lateral collision with road users originating from a different approach. This component is part of a larger system relying on video sensors to provide queue lengths and spatial occupancy that are used for real time traffic control and monitoring. The method is evaluated on data collected during a real world experiment. 

\section{Introduction}
Collisions at intersections make up a high proportion of total collisions all over the world, 39.7~\% of all collisions and 21.8~\% of fatal collisions in the US for instance \citep{nhtsa08tsf}. Traffic signals are installed at intersections according to warrants on traffic volumes and safety. Over the years, there has been considerable interest in improving signalized intersections to minimize delays and stops. It has become possible to develop traffic control strategies that can adapt to current traffic conditions in real-time thanks to the advancement of sensors. However not much is known about the impact on safety of traffic control strategies, particularly adaptive ones. 

The CRONOS strategy is a real-time adaptive traffic control strategy developed in the 1990s~\citep{boillot06real-time}. It relies on video sensors that provide queue lengths and spatial occupancy rates every second~\citep{aubert96automatic}. Its algorithm controls the traffic lights in order to minimize total delay over a short-range time horizon. The CRONOS strategy was compared to a standard time-plan based control strategy with vehicle-actuated ranges. During the real world experiment, the two strategies alternately controlled a real intersection in the suburbs of Paris over several months. 
The traffic databases recorded during the CRONOS strategy assessment enable us to obtain quantitative results in a real world setting and to compare the exposure between strategies and between traffic volume conditions. 

An original model was developed to evaluate the exposure to lateral collision in intersections. Our model considers the conditions in which road users go through an intersection with respect to the presence of road users in cross-traffic approaches. This paper describes the method used to estimate automatically this measure of exposure using occupancy data provided by the video sensors, which is the same data that supplies the CRONOS strategy. The authors are referred to~\citep{saunier05thesis} for more details on the background of this work. 
The outline of the paper is as follows: section~\ref{sec:expos-later-coll} defines the exposure to lateral collision used in this work, section~\ref{sec:data-description} presents the data, section~\ref{sec:rule-based-method} describes the method and its evaluation, before the conclusion in section~\ref{sec:conclusion}.

\section{The Exposure to Lateral Collision in Signalized Intersections}\label{sec:expos-later-coll}

The concept of exposure to collision has been introduced to ``take account of the amount of opportunity for collisions which the driver of the traffic system experiences'' \citep{chapman73exposure}. \citet{archer04methods} defines exposure as a ``measure of spatial or temporal duration in the traffic system in relation to the number of dynamic system objects, road users, vehicles, etc''. In its most general definition, any necessary situation for a collision to occur can be considered as exposure to collision. 

The presence of road users on the cross-traffic approach is a necessary condition to the occurrence of a lateral collision in the conflict zone: while there is no risk of lateral collision if the cross-traffic approach is empty, this risk is non-zero if there is at least one road user in the cross-traffic approach. This latter situation is called a \emph{critical situation} for lateral collisions. In signalized intersections, red-light running is a second necessary condition for a collision to happen in the conflict zone that will not be investigated in this work. 

\begin{figure}[htb!]
  \begin{center}
    \includegraphics[width=0.6\textwidth]{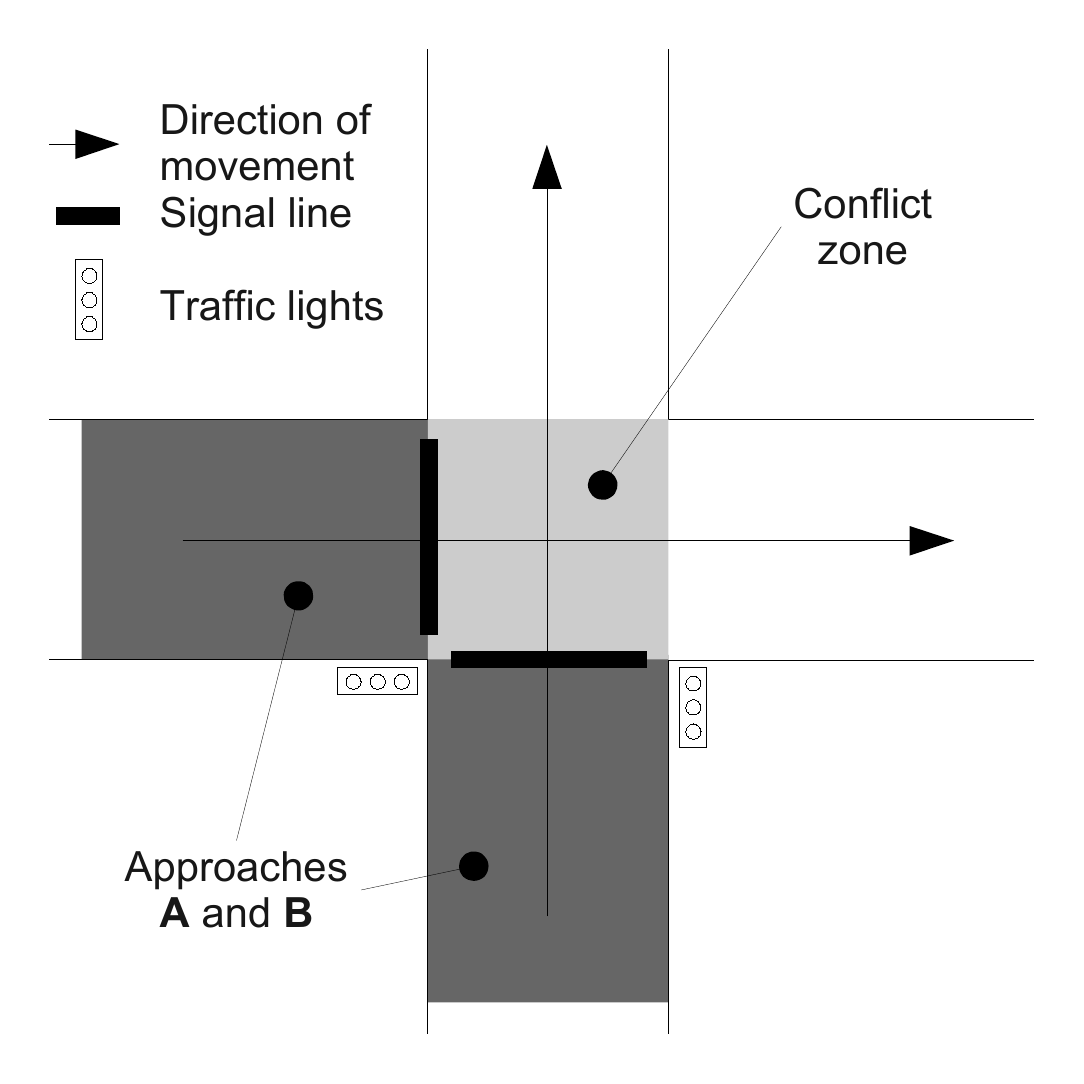}
    \caption{Simple intersection of two one-way roads.}
    \label{fig:simple-intersection}
  \end{center}
\end{figure}

\begin{figure}[htb!]
  \begin{center}
    \includegraphics[]{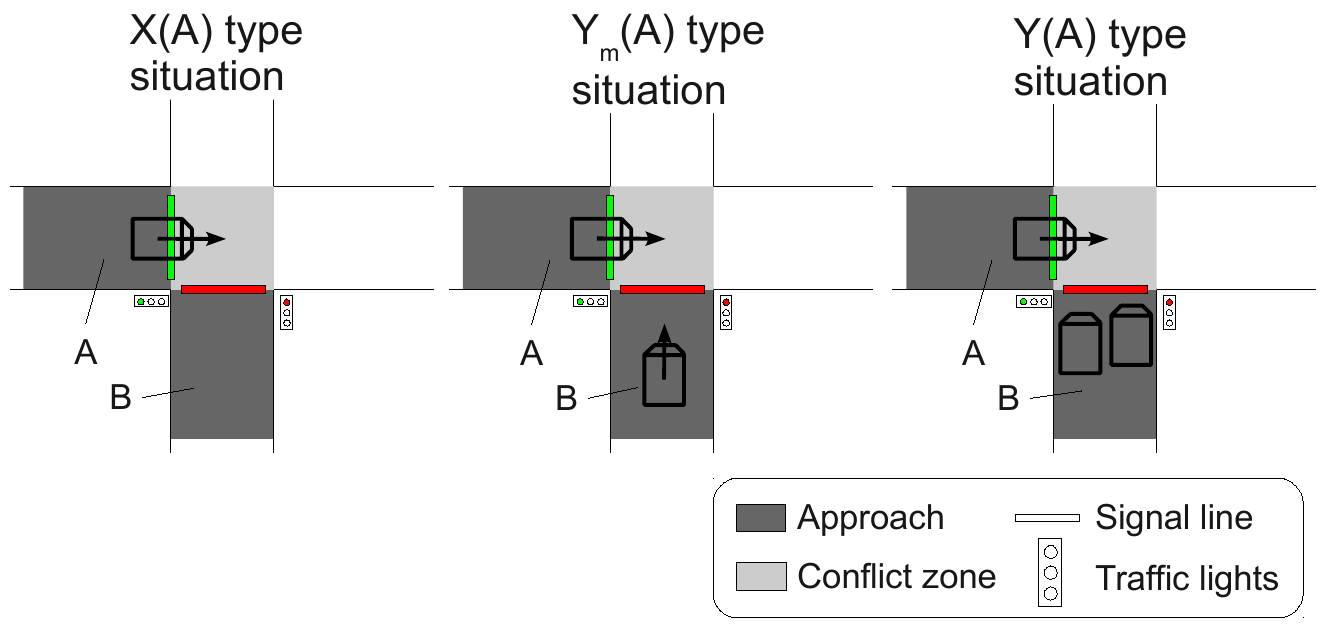}
    \caption{The different crossing situations for stream~A.}
    \label{fig:exposure-types}
  \end{center}
\end{figure}

The exposure to lateral collision in intersections is defined as the temporal duration of critical situations, which occur when the stream that is given right-of-way by the signal goes through the conflict zone while there is at least one road user in the cross-traffic approach. 
In the simple case of a signalized intersection of two one-way roads (see Figure~\ref{fig:simple-intersection}), a road user enters the conflict zone from one of the two approaches, denoted A and~B. Let us consider a given period $T$ and one of the two traffic streams originating from the two approaches, say A. We adopt the following definitions, illustrated in Figure~\ref{fig:exposure-types}. 

\begin{itemize}
\item Duration $Z(A)$ is the cumulated amount of time during which road users of stream A are crossing the conflict zone within period T.
\end{itemize}

When road users of stream A are crossing the conflict zone, there are two possible situations: absence or presence of road users in the cross-traffic approah B.

\begin{itemize}
\item Duration $X(A)$ is the cumulated amount of time during which road users of stream A are crossing the conflict zone when B is empty. 
\item Duration $Y(A)$ is the cumulated amount of time during which road users of stream A are crossing the conflict zone when road users are present in~B.
\end{itemize}

It follows that: $Z(A) = X(A) + Y(A)$.
 
$Y(A)$ can be further sub-divided by considering whether all the close road users in B are stationary or at least one close road user is moving in B (Figure~\ref{fig:exposure-types}). A close road user is a road user in the cross traffic approach such that there is no other road user between itself and the conflict zone, e.g.\ the first road user in each lane of an entry cross traffic approach. $Y_m(A)$ is the cumulated amount of time during which road users of stream A are crossing the conflict zone when at least one close road user is moving in B, with $Y_m(A) \leq Y(A)$. The definitions are symmetric with respect to A and B.

\section{Data Description}\label{sec:data-description}
The system used in this work extracts spatial occupancy measurements of the whole intersection from video sensors. It can be seen as a virtual grid of sensors covering the surface of the intersection. The coverage is bi-dimensional for the inner part of the intersection, and linear for each lane of the approaches. Each grid unit may take one of six possible states representing the dynamics over one second of the road users that passed over the corresponding part of the roadway (see Figure~\ref{fig:situation-detection}). The six possible states can be aggregated into four main states: emptiness, moving presence, stationary presence and end of presence. 

The data is difficult to interpret at grid-unit level. Since the spatio-temporal definition is limited, it is not possible to identify individual road users and it is difficult to estimate distances between them. To make interpretations more robust to noise, components of neighboring grid units are considered. Precisely, the neighbours of a grid unit are the four (or two for lane approaches) closest units, i.e.\ the ones that are vertically and/or horizontally adjacent. %
The connected components are formed independently for each grid-unit state in each functional zone of the intersection. The connected components correspond to either moving or stationary groups of road users and will be referred to in this way from now on. Thresholds are further applied to the number of units in groups below which the groups are discarded (more illustrations are provided in \citep{saunier03automatic-detection}): they were determined empirically to be 2 and 3 units respectively for moving and stationary groups of road users. 

\section{A Rule-based Method for the Detection of Critical Situations}\label{sec:rule-based-method}
\subsection{Detecting Stop Line Crossings}\label{detecting-stop-line}
The first step to detect critical situations and measure the exposure to lateral collision is to identify the origin of moving groups of road users detected in the conflict zone. This is achieved by detecting stop line crossings, using a method developed previously by one of the authors \citep{midenet98som-nimes}. It relies on rules counting the number of grid units in some states on both sides of the stop line in order to determine the situations upstream and downstream of the stop line. This is a generic method that allows to detect crossings from one side to the other. 

Two subsets of grid units, respectively upstream and downstream of the stop line, i.e.\ in the storage and conflict zones, are delimited, and henceforth called only ``upstream'' and ``downstream''. 
The number of grid units taken into account for each zone may be adjusted in the system. %
For each set of grid units, the ratio $\tau_{state}$ is defined as the number of grid units in the given state, denoted $n_{state}$, divided by the total number of grid units in the set. 
To make the presentation of the rules simpler, two new grid unit states are built upon the four possible values:

\begin{description}
\item[Presence] corresponds to moving presence, end of presence or stationary presence
\item[Movement] corresponds to moving presence or end of presence
\end{description}

The ``movement state'' of each subset of grid units is set as either $Empty$, $Stationary$, $Past\ Movement$, $Future\ movement$ or $Movement$ according to the rules, presented as algorithm~\ref{alg:regles-aval} for the downstream zone and algorithm~\ref{alg:regles-amont} for the upstream zone. For these states to be taken into account, the downstream and upstream zones must meet the crossing conditions described in algorithm~\ref{alg:regles-zones}, and a crossing between the two zones is detected according to the rule of algorithm~\ref{alg:regles-ligne}. 

\begin{myalgorithm}{Qualification rules for a downstream zone.}
  \begin{tabbing}
    \keyw{if} $(\tau_{presence} < 10\ \%)$ and $(n_{moving\ presence} = 0)$\\
    \quad \= the state of downstream zone is $Empty$\\
    \keyw{if} $(\tau_{presence} \geq 10\ \%)$ and $(\tau_{movement} < 10\ \%)$ et $(n_{moving\ presence} = 0)$\\
    \quad the state of the downstream zone is $Stationary$\\
    \keyw{if} the state of the downstream zone is neither $Empty$ nor $Stationary$\+ \\
    \keyw{if} $n_{movement} = n_{end\ of\ presence}$\\
    \quad the state of the downstream zone is $Past\ Movement$\- \\
    \keyw{else}\\
    \quad the state of the downstream zone is $Movement$\\
  \end{tabbing}
  \label{alg:regles-aval}
\end{myalgorithm}

\begin{myalgorithm}{Qualification rules for an upstream zone.}
  \begin{tabbing}
    \keyw{if} $(\tau_{presence} < 10\ \%)$ and $(n_{end\ of\ presence} = 0)$\\
    \quad \= the state of downstream zone is $Empty$\\
    \keyw{if} $(\tau_{presence} \geq 10\ \%)$ and $(\tau_{movement} < 10\ \%)$ et $(n_{end\ of\ presence} = 0)$\\
    \quad the state of the upstream zone is $Stationary$\\
    \keyw{if} the state of the upstream zone is neither $Empty$ nor $Stationary$\+ \\
    \keyw{if} $n_{movement} = n_{moving\ presence}$\\
    \quad the state of the downstream zone is $Future\ Movement$\- \\
    \keyw{else}\\
    \quad the state of the downstream zone is $Movement$\\
  \end{tabbing}
  \label{alg:regles-amont}
\end{myalgorithm}

\begin{myalgorithm}{Rules determining if the crossing conditions are met for upstream and downstream zones ($\neg Movement$ means any zone state except $Movement$).}
  \begin{tabbing}
    \keyw{if} the state of the upstream zone is $Movement$\\
    \quad the crossing condition for the upstream zone is $met$\\
    \keyw{if} the state of the upstream zone changes from $Movement$ to $\neg Movement$\\
    \quad the crossing condition for the upstream zone is $not\ met$\\
    \keyw{if} (the state of the downstream zone is $Movement$) and (the crossing condition of the upstream zone is $met$)\\
    \quad the crossing condition for the downstream zone is $met$\\
    \keyw{else}\\
    \quad the crossing condition for the downstream zone is $not\ met$\\
  \end{tabbing}
  \label{alg:regles-zones}
\end{myalgorithm}

\begin{myalgorithm}{Rule to detect a crossing between an upstream and a downstream zone.}
  \begin{tabbing}
    \keyw{if} the crossing conditions on the upstream and downstream zones are $met$\\
    \quad \keyw{if} (the state of the upstream zone is $Movement$) and ($n_{movement} \geq 0.5\ n_{presence}$)\\
    \quad \quad A crossing between the two zones is detected\\
  \end{tabbing}
  \label{alg:regles-ligne}
\end{myalgorithm}

Finally, a decision is made about the origin of the moving groups of road users in the conflict zone. 
It is assumed that all road users newly detected in a conflict zone originate from the same approach and crossed the stop line during the last second. 
It is simple if the method described above detected only one crossing from an upstream zone. If crossings were detected from the two upstream zones, the origin of the road users detected in the conflict zone is the upstream zone for which the traffic light was not red (if the two upstream traffic lights are red, the origin is the upstream zone for which the traffic light was most recently green).

\subsection{The Exposure Detection Method}
At each second, in each zone of the intersection, groups of road users are detected, the origin of moving groups in conflict zones is identified, and the durations $Z$, $X$, $Y$ and $Y_m$ are updated following the algorithm~\ref{alg:detection-method}. 

\begin{myalgorithm}{The method for the detection of exposure.}
  \begin{tabbing}
    \keyw{if} at least one group of road users is crossing the conflict zone from approach A (respectively B)\\
    \quad $Z(A)$ (respectively $Z(B)$) is incremented\\
    \quad \keyw{if} there is at least one group of road users in the cross-traffic approach~B (respectively A)\\
    \quad \quad $Y(A)$ (respectively $Y(B)$) is incremented\\
    \quad \quad \keyw{if} one of the close groups of road users in B (respectively A) is moving\\
    \quad \quad \quad $Y_{m}(A)$ (respectively $Y_{m}(B)$) is incremented\\
    \quad \keyw{else}\\
    \quad \quad $X(A)$ (respectively $X(B)$) is incremented\\
  \end{tabbing}
  \label{alg:detection-method}
\end{myalgorithm}

\begin{figure}[htb!]
  \begin{center}
    \includegraphics[width=\textwidth]{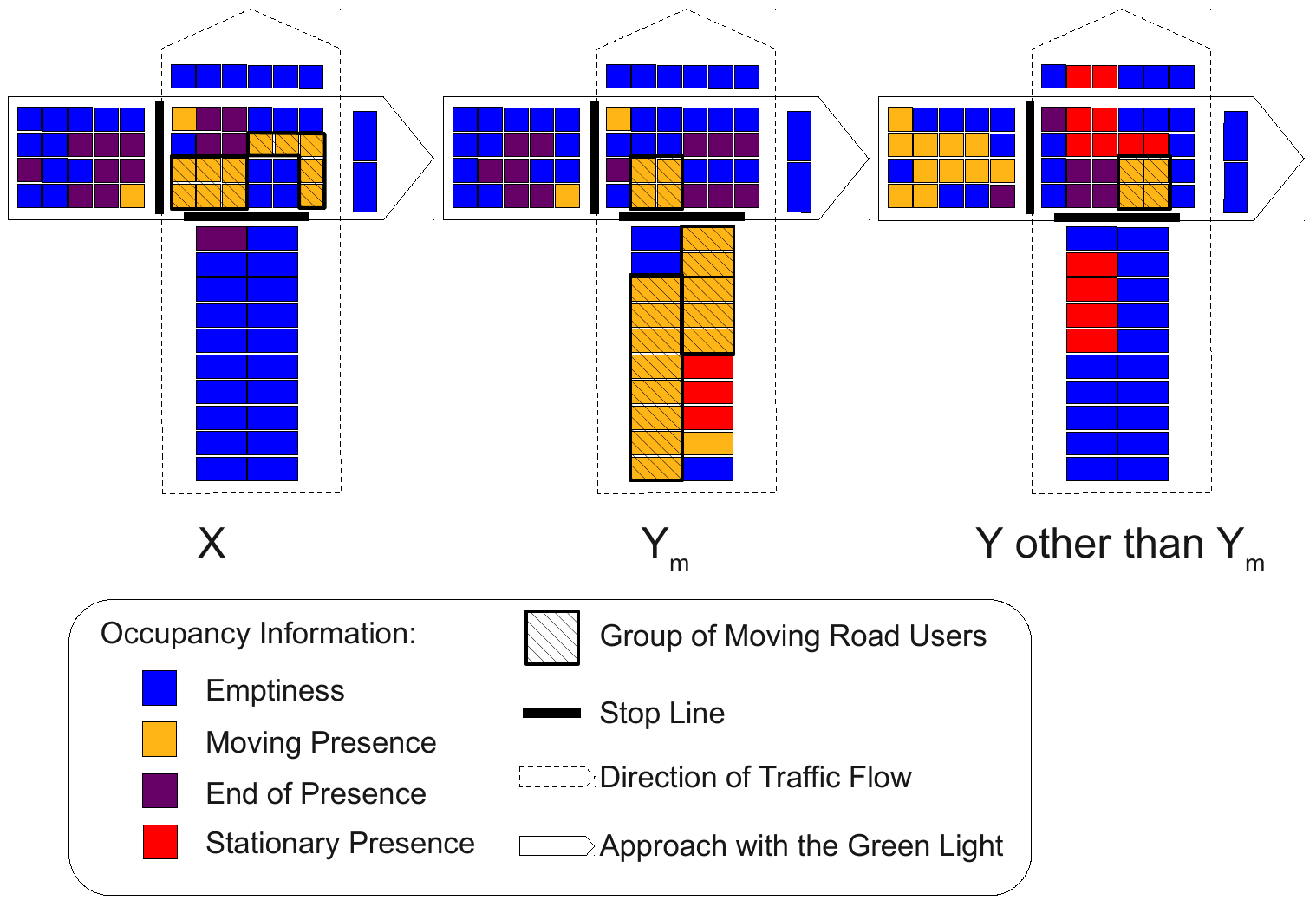}  
    \caption{Examples of the data collected and the detection of critical situations.}
  \label{fig:situation-detection}
  \end{center}
\end{figure}

In each example in Figure~\ref{fig:situation-detection}, a group of moving road users is detected in the conflict zone (corresponding to the connected components constituted by 6 and 4 grid units in state ``Moving Presence''). In the middle and right examples, groups of road users are detected in the cross-traffic approach: a critical situation is thus identified and $Y$ is incremented. Some close road users in the cross-traffic approach are moving in the middle example, hence it is counted as $Y_m$, while all road users in the cross-traffic approach are stationary in the right example. %

\subsection{Evaluation}
The detection rules were evaluated on a subset of the data for two sub-intersections of the experimental site and the two traffic control strategies. Since the duration $Y_m$ is a small subset of $Y$, more time is necessary to collect enough data for $Y_m$. Using the video recordings, 10 and 20~minutes of data in peak traffic conditions were manually annotated respectively for $Y$ and $Y_m$, for each strategy. The validation results are presented as a binary classification problem for durations $Y$ and $Y_m$  independently in Table~\ref{tab:detection-results}. It appears that automated detection of critical interactions is reliable. The results are better for $Y$, where recall reaches more than 92~\% for three intersection/origins, than for $Y_m$, where recall is superior to 75~\%. Moreover the performance of the method is fairly similar for both traffic control strategies.

\begin{table}[htb!]
  \begin{center}
    \caption{Evaluation results for the detection of instants of critical interactions $Y$ (upper half) and $Y_m$ (lower half). The results are presented in a binary classification framework, with the instants of interactions being the positive class. True positives $TP$ are the instants of critical interaction detected as such, false positives $FP$ are the instants without any critical interaction detected as critical interaction, and false negatives $FN$ are the instants of critical interactions that are not detected. The performance measures $Precision = TP/(TP+FP)$ and $Recall = TP/(TP+FN)$ are computed.}
  \label{tab:detection-results}
  \begin{tabular}{llllllll}
    \hline
    &&  & TP & FP & FN & Recall & Precision\\
    \hline
    $Y$&baseline&E	&105	&6	&32	&77~\%	&95~\%\\
    &	&s	&200	&29	&18	&92~\%	&87~\%\\
    &	&N	&104	&28	&7	&94~\%	&79~\%\\
    &	&e	&110	&13	&6	&95~\%	&89~\%\\
    \hline
    &CRONOS&E	&53	&2	&14	&79~\%	&96~\%\\
    &	&s	&198	&14	&12	&94~\%	&93~\%\\
    &	&N	&135	&43	&6	&96~\%	&76~\%\\
    &	&e	&89	&27	&9	&91~\%	&77~\%\\
    \hline
    $Y_m$& baseline & E & 45 & 2 & 12 & 79~\% & 96~\%\\
    && s & 20 & 15 & 6 & 77~\% & 57~\%\\
    && N & 36 & 12 & 2 & 95~\% & 75~\%\\
    && e & 18 & 5 & 0 & 100~\% & 78~\%\\
    \hline
    &CRONOS & E & 24 & 1 & 7 & 77~\% & 96~\%\\
    && s & 15 & 10 & 5 & 75~\% & 60~\%\\
    && N & 31 & 12 & 2 & 94~\% & 72~\%\\
    && e & 15 & 5 & 3 & 83~\% & 75~\%\\
    \hline
  \end{tabular}
\end{center}
\end{table}

\section{Conclusion}\label{sec:conclusion}
This paper has presented a method to detect automatically the exposure to lateral collision in an intersection. It relies on occupancy information extracted from video data. The approach is accurate enough to be used to process large amounts of data automatically. The result is further analyzed to estimate severity indicators related to road users' speeds \citep{saunier04stream-based-learning,saunier10creating} and to compare the influence of the traffic control strategies on the exposure to lateral collision \citep{saunier05thesis,midenet10exposure}. 

A fair question is the genericity of the method. The model of exposure may be applied to all intersections, signalized or not, but was developed in the particular case of signalized intersections with protected left turns. The particular system and the data upon which the method relies may not be easily replicated. However, the method was described as completely as possible so that it can serve as inspiration for other work. 

\bibliographystyle{plainnat}
\bibliography{references,saunier}

\begin{thebibliography}{11}
\providecommand{\natexlab}[1]{#1}
\providecommand{\url}[1]{\texttt{#1}}
\expandafter\ifx\csname urlstyle\endcsname\relax
  \providecommand{\doi}[1]{doi: #1}\else
  \providecommand{\doi}{doi: \begingroup \urlstyle{rm}\Url}\fi

\bibitem[Archer(2004)]{archer04methods}
J.~Archer.
\newblock \emph{Methods for the Assessment and Prediction of Traffic Safety at
  Urban Intersections and their Application in Micro-simulation Modelling}.
\newblock Academic thesis, Royal Institute of Technology, Stockholm, Sweden,
  December 2004.
\newblock URL \url{http://urn.kb.se/resolve?urn=urn:nbn:se:kth:diva-143}.

\bibitem[Aubert et~al.(1996)Aubert, Bouzar, Lenoir, and
  Blosseville]{aubert96automatic}
D.~Aubert, S.~Bouzar, F.~Lenoir, and J.-M. Blosseville.
\newblock Automatic vehicle queue measurement at intersections using
  image-processing.
\newblock In \emph{8$^{th}$ International Conference on Road Traffic Monitoring
  \& Control}, volume 422, pages 100--104, London, 1996.

\bibitem[Boillot et~al.(2006)Boillot, Midenet, and
  Pierrel\'ee]{boillot06real-time}
F.~Boillot, S.~Midenet, and J.-C. Pierrel\'ee.
\newblock The real-time urban traffic control system cronos: Algorithm and
  experiments.
\newblock \emph{Transportation Research Part C: Emerging Technologies},
  14\penalty0 (1):\penalty0 18--38, 2006.
\newblock ISSN 0968-090X.
\newblock \doi{10.1016/j.trc.2006.05.001}.

\bibitem[Chapman(1973)]{chapman73exposure}
R.~Chapman.
\newblock The concept of exposure.
\newblock \emph{Accident Analysis \& Prevention}, 5\penalty0 (2):\penalty0
  95--110, June 1973.

\bibitem[Midenet(1998)]{midenet98som-nimes}
S.~Midenet.
\newblock Cartes auto-organisatrices pour l'interpr\'etation de mesures
  spatiales et la description du trafic au centre de carrefours.
\newblock In \emph{Session Transport du congr\`es ``Syst\`emes complexes,
  syst\`emes intelligents et interface''}, volume 134-135-136, pages 240--244,
  N\^{\i}mes, May 1998. La lettre de l'IA.

\bibitem[Midenet et~al.(2010)Midenet, Saunier, and Boillot]{midenet10exposure}
S.~Midenet, N.~Saunier, and F.~Boillot.
\newblock Exposure to lateral collision in signalized intersections with
  different traffic control strategies.
\newblock \emph{Accident Analysis \& Prevention}, 2010.
\newblock Submitted.

\bibitem[{NHTSA}(2008)]{nhtsa08tsf}
{NHTSA}.
\newblock Traffic safety facts.
\newblock National Highway Traffic Safety Administration, 2008.
\newblock URL \url{http://www-nrd.nhtsa.dot.gov/Pubs/811170.pdf}.

\bibitem[Saunier(2005)]{saunier05thesis}
N.~Saunier.
\newblock \emph{{I}ncidence de la régulation d'un carrefour à feux sur le
  risque des usagers. {A}pprentissage d'indicateurs par sélection de données
  dans un flux}.
\newblock PhD thesis, ENST Paris, June 2005.
\newblock URL \url{http://tel.archives-ouvertes.fr/tel-00009447/fr/}.
\newblock {ENST} 2005 E 011.

\bibitem[Saunier and Midenet(2010)]{saunier10creating}
N.~Saunier and S.~Midenet.
\newblock Creating ensemble classifiers through data selection and order;
  application to the online learning of road safety indicators.
\newblock \emph{Pattern Analysis \& Applications}, 2010.
\newblock Submitted.

\bibitem[Saunier et~al.(2003)Saunier, Midenet, and
  Grumbach]{saunier03automatic-detection}
N.~Saunier, S.~Midenet, and A.~Grumbach.
\newblock {A}utomatic detection of vehicle interactions in a signalized
  intersection.
\newblock In \emph{$16^{th}$ International Cooperation on Theories and Concepts
  in Traffic Safety Workshop}, Soesterberg, The Netherlands, October 2003.
\newblock URL \url{http://www.ictct.org/Workshops/03-Soesterberg/Saunier.pdf}.

\bibitem[Saunier et~al.(2004)Saunier, Midenet, and
  Grumbach]{saunier04stream-based-learning}
N.~Saunier, S.~Midenet, and A.~Grumbach.
\newblock {S}tream-based {L}earning through {D}ata {S}election in a {R}oad
  {S}afety {A}pplication.
\newblock In E.~Onaindia and S.~Staab, editors, \emph{STAIRS 2004, Proceedings
  of the Second Starting AI Researchers' Symposium}, volume 109 of
  \emph{Frontiers in Artificial Intelligence and Applications}, pages 107--117,
  Valencia, Spain, August 2004. IOS Press.
\newblock ISBN 1 58603 451 0.

\end{thebibliography}
\end{document}